\documentclass[letterpaper, 10 pt, conference]{ieeeconf}  % Comment this line out if you need a4paper

\IEEEoverridecommandlockouts                              % This command is only needed if 
\usepackage{graphics} % for pdf, bitmapped graphics files
\usepackage{float}
\usepackage{epsfig} % for postscript graphics files
\usepackage{mathptmx} % assumes new font selection scheme installed
\usepackage{times} % assumes new font selection scheme installed
\usepackage{amsmath} % assumes amsmath package installed
\usepackage{amssymb}  % assumes amsmath package installed
\usepackage{subcaption}
\usepackage{algorithm}
\usepackage[noend]{algpseudocode}
\usepackage{hyperref}
\hypersetup{
    colorlinks=true,
    linkcolor=blue,
    filecolor=magenta,      
    urlcolor=cyan,
}

\title{\LARGE \bf
Real-Time Ground-Plane Refined LiDAR SLAM
}

\author{
    Fan Yang\\
    Robotics Institute\\
    Carnegie Mellon University\\
    \texttt{fanyang2@andrew.cmu.edu}
  \and
    Mengqing Jiang\\
    Robotics Institute\\
    Carnegie Mellon University\\
    \texttt{mengqinj@andrew.cmu.edu}
    \and
    Chenxi Xu\\
    Robotics Institute\\
    Carnegie Mellon University\\
    \texttt{chenxix@andrew.cmu.edu}
}

\begin{document}

\maketitle
\thispagestyle{empty}
\pagestyle{empty}

%%%%%%%%%%%%%%%%%%%%%%%%%%%%%%%%%%%%%%%%%%%%%%%%%%%%%%%%%%%%%%%%%%%%%%%%%%%%%%%%
\begin{abstract}

SLAM system using only point cloud has been proven successful in recent years, such as~\cite{ pfrunder2017real, shan2018lego, tang2018learning, zhang2014loam}. In most of these systems, they extract features for tracking after ground removal, which causes large variance on the z-axis. Ground actually provides robust information to obtain $\left[t_z, \theta_{roll}, \theta_{pitch}\right]$. In this project, we followed the LeGO-LOAM~\cite{shan2018lego}, a light-weighted real-time SLAM system that extracts and registers ground as an addition to the original LOAM~\cite{zhang2014loam}, and we proposed a new clustering-based method to refine the planar extraction algorithm for ground such that the system can handle much more noisy or dynamic environments. We implemented this method and compared it with LeGo-LOAM on our collected data of CMU campus, as well as a collected dataset for ATV (All-Terrain Vehicle) for off-road self-driving. Both visualization and evaluation results show obvious improvement of our algorithm.

\end{abstract}

%%%%%%%%%%%%%%%%%%%%%%%%%%%%%%%%%%%%%%%%%%%%%%%%%%%%%%%%%%%%%%%%%%%%%%%%%%%%%%%%
\section{INTRODUCTION}

SLAM algorithm relies on sensor data to achieve accurate pose estimation. Especially, LiDAR is commonly used to efficiently make high-resolution 3D maps, which is essential for the localization of autonomous driving robots and 3D reconstruction.
When using LiDAR as the sensor, what we get is a set of point clouds, and would require frame-to-frame registration to acquire relative transformation between frames. The registration step is a challenging problem and remains as a bottleneck for real-time localization. 

Most of the current registration approaches are based on iterative closest point (ICP) registration. The ICP algorithm aligns two sets of points iteratively, where each iteration consists of first finding the best point-to-point correspondences, and then computing a rigid transformation between all corresponding pairs of points. 
ICP works well for point cloud registration, but it is not the best when dealing with the registration of scene-level point cloud. It only leverages point-wise distance to find the optimized correspondence, in the sense that global information is being ignored and, obviously, the optimization takes a very long time to converge, making it hard to do real-time processing.

The problem becomes how to extract the planar feature and leverage it in point cloud registration. As we all know, the largest plane in real-world 3D scan is the ground. In most SLAM algorithms, ground removal is the very first step since it provides no obvious feature for tracking. However, the registration on point cloud without ground has a large variance on the z-axis, especially under the off-road circumstance. LeGO-LOAM~\cite{shan2018lego} uses this idea and implemented it based on the real-time SLAM system, LOAM~\cite{zhang2014loam}. It achieved very impressive and robust SLAM results in the wild but failed when the LiDAR sensor is not vertical to the ground or the extracted ground contains too much noise.

In this work, we propose to apply ground finding algorithms in the registration process, and use a new cluster-based ground extraction algorithm, compared to LEGO-LOAM. The ground feature will benefit the pose estimation on the z-axis, as well as roll and pitch angles, which could act as a supplement of ICP registration. We collect our own LiDAR data on CMU campus (both outdoor and indoor), and compare the experimental results of our algorithm and LEGO-LOAM on those data. We further evaluate our algorithm on a collected off-road environment data from an HDL-64E LiDAR and beat LEGO-LOAM in final drift.

In the following sections, we will first introduce several previous works related to our project in Section~\ref{sec:related-work}, and then discuss our proposed methods and collected dataset in Section~\ref{sec:methods} and Section~\ref{sec:data} respectively. Section~\ref{sec:implement} details our implementation and Section~\ref{sec:experiments} shows the experiment results. At last, we draw to conclusion in Section~\ref{sec:conclusion}.

\section{RELATED WORK}
\label{sec:related-work}

 \begin{figure*}[thpb]
    \centering
    
    \begin{subfigure}[b]{0.33\textwidth}
        \centering
        \vskip 0pt
        \includegraphics[width=.95\textwidth]{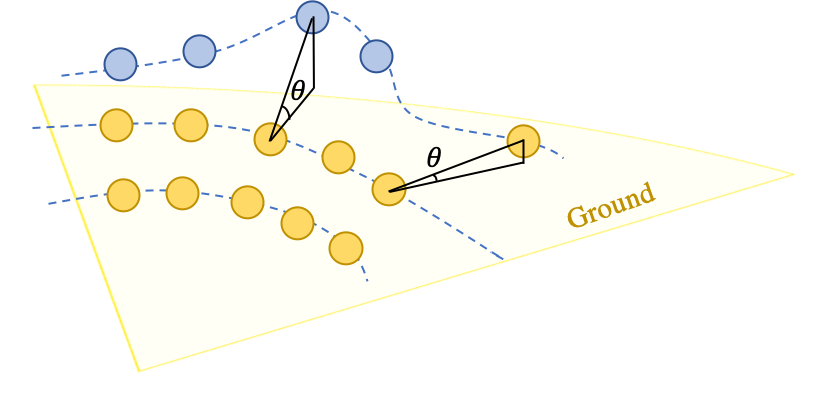}
      \caption{Ground extraction by LEGO-LOAM. \\
      Yellow dots are estimated ground points and blue dots are estimated non-ground points. $\theta$ is the vertical angle.}
      \label{related-lego}
    \end{subfigure}%
    ~ 
    \begin{subfigure}[b]{0.33\textwidth}
        \centering
        \vskip 0pt
        \includegraphics[width=.95\textwidth]{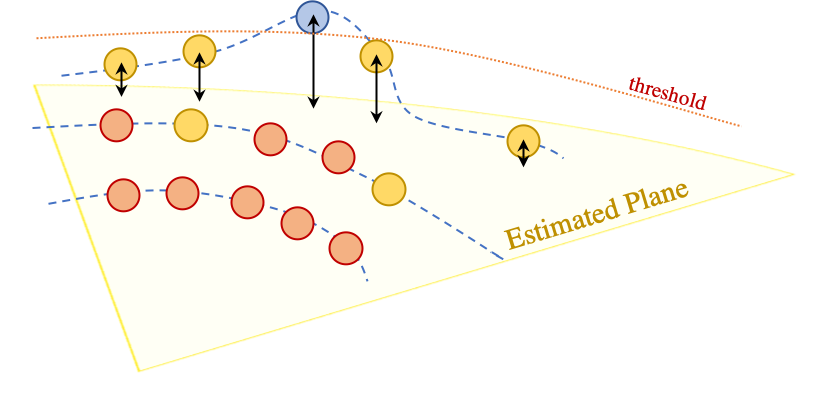}
      \caption{Ground Plane Fitting. \\
      Yellow dots are estimated ground points, blue dots are estimated non-ground points and red dots are Lowest Point Representatives.}
      \label{related-gpf}
    \end{subfigure}%
    ~ 
    \begin{subfigure}[b]{0.33\textwidth}
        \centering
        \vskip 0pt
        \includegraphics[width=.95\textwidth]{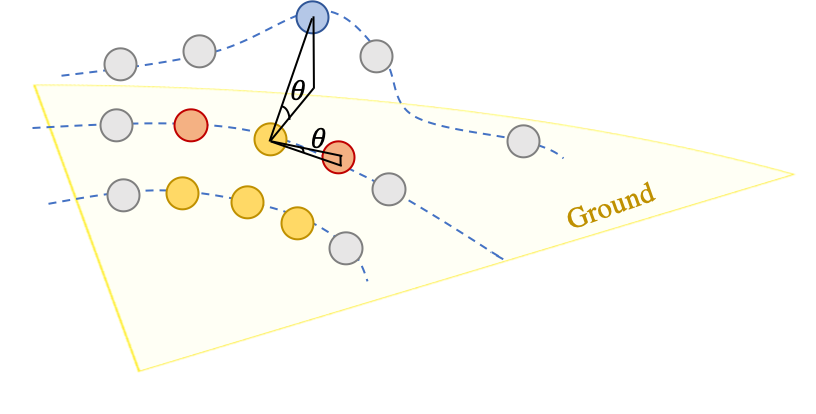}
      \caption{Our ground extraction method. \\
      Yellow dots are estimated ground points, blue dots are estimated non-ground points, red dots are newly labelled ground points and gray dots are unprocessed points. }
      \label{ours}
    \end{subfigure}%
    ~ 
    \caption{Comparison of three ground extraction algorithms.}
\end{figure*}
We first conducted a literature review on successful SLAM systems in recent years, and then we investigated those methods that are using the planar features for registration. At last, we looked into the papers related to ground extraction specifically.

LOAM~\cite{zhang2014loam} proposed a real-time system that performs odometry at a high frequency but low fidelity to estimate the velocity of the lidar and, at the same time, matches and registers the point cloud to create a map at a much lower frequency in parallel. Their main goal is to perform SLAM in real-time and minimize drift without loop closure. Another big challenge of this work is that the moving and rotating lidar scanner can make it very difficult to estimate motion and map point cloud. This paper addressed those problems by taking advantage of both scan-to-scan registration and scan-to-map registration. The key idea is to perfectly combine the algorithm with high frequency but low fidelity and the algorithm with low frequency but high fidelity. Additionally, features are extracted as edge points and planar points so as to further reduce the time of computation.

LEGO-LOAM~\cite{shan2018lego} is an extension of section 2.1 of LOAM. This algorithm is implemented for unmanned ground vehicles(UGVs). The paper proposed a lightweight LOAM to work on a small-scale embedded system and a ground-optimized model to account for the sensor noise from the ground. To optimize on-ground registration, they introduce a two-step optimization for pose estimation. First they extract planar features from the ground to obtain $\left[t_z, \theta_{roll}, \theta_{pitch}\right]$; Second, they match edge features extracted from the segmented point cloud to obtain the other three transformations$\left[t_x, t_y, \theta_{yaw}\right]$. With these updates, LeGO-LOAM achieves similar or better accuracy with a reduced computational expense.

Additionally, we looked into how LEGO-LOAM extracts the ground points. As how the code is implemented, the ground points are extracted by the following steps, also as visualized in Figure~\ref{related-lego}:

\begin{enumerate}
    \item Assign vertical index and horizontal index for each point from the point cloud scan, such that the points on one row or on one column can be easily retrieved.
    \item Check through all points in the half lower part of the scan. For example, use 7 lowest rows for VLP-16, or 15 lowest rows for HDL-32E LiDAR.
    \item Compute vertical angle between two adjacent points at the same column and label both points as ground points if the vertical angle is smaller than 10 degrees.
\end{enumerate}

This algorithm fails when there is too much noise, especially for points on some surfaces other than the ground. For example, if the LiDAR scanning indoor is not vertical to the ground, the scan will contain the points of the surface not only from the ground, but also from the ceiling or tables as well, and the algorithm could fail to separate them and label them all as ground.

Some other LiDAR segmenters libraries for ROS\footnote{https://github.com/LidarPerception/segmenters\_lib} implemented the Ground Plane Fitting algorithm from~\cite{zermas2017fast}. As shown in Figure~\ref{related-gpf}, \cite{zermas2017fast} finds the lowest point representative (LPR) as the initial ground seeds and fit a plane model. Then each point is evaluated against the estimated plane model. If the distance from the point to the plane is lower than the manually defined threshold, this point is labeled as ground. This method is very efficient, however, this algorithm is very sensitive to the threshold and number of LPR, which not only makes it very hard to do parameter tuning but also makes the extracted ground points contain much noise. In our experiments, we found that the extracted ground contains many points of walls in our indoor scene data, which leads to the wrong estimation of the z-value of the ground.

\section{METHODS}
\label{sec:methods}
\subsection{Ground Point Clustering and Extraction}
Inspired by the cloud segmentation method in LeGO-LOAM, we proposed a new method for ground point extraction. The basic idea is to cluster points that have a relatively low angle between adjacent points.

Specifically, we use all points on the lowest $frac{1}{4}$ part of the scan as initial seeds (i.e. 4 rows for VLP-16 or 8 rows for HDL-32E LiDAR), and then cluster points from near to far according to neighborhood vertical angles, as visualized in Figure~\ref{ours}. The clusters whose number of points is over the threshold are labeled as ground. The pseudo-code is given in Algorithm~\ref{algorithm}.

\begin{algorithm}
\caption{Our proposed ground extraction algorithm}\label{algorithm}
\begin{algorithmic}[1]
\Procedure{GroundExtraction}{}
\For{$j \gets 0 $ to \textit{Horizon\_SCAN}}
    \For{$i \gets 0 $ to \textit{groundScanInd}}
        \State \Call{GroundCluster}{$i$, $j$}
    \EndFor
\EndFor
\EndProcedure

\Procedure{GroundCluster}{$row$, $col$}
\State $ q \gets $ New Queue
\State $cluster \gets$ New Array 
\State push ($row$, $col$) to $q$ 
\While{$q$ is not empty}
    \State $P \gets $ top of $q$
    \State pop $q$
    \For{$P_i \gets$ neighboors of $P$}
        \State $\theta \gets$ vertical angle between $P$ and $P_i$
        \If{$\theta < $ \textit{angleThreshold}}
            \State push $P_i$ to $q$
            \State add $P_i$ to $cluster$
        \EndIf
    \EndFor
    \If{size of $cluster >$ \textit{sizeThreshold}}
        \State label points in $cluster$ as ground
    \EndIf
\EndWhile

\EndProcedure
\end{algorithmic}
\end{algorithm}

\subsection{Point Clustering}
After ground segmentation, another step is point clustering(or point segmentation). Doing point segmentation can help ease the time complexity of feature matching and increase feature extraction accuracy. Another advantage is filtering out small clusters that are noisy and unreliable. This method is taken from LeGO-LOAM\cite{shan2018lego}.

The basic idea of point segmentation is to assign the same label to points that are relatively close to each other. To filter out unreliable features, like tree leaves in the outdoor scenes, clusters with less than 30 points will be removed. In the feature matching process, features of the current frame only need to be compared with points with the same cluster label.

After this process, only features that represent large objects, like walls, door frames, table, or chair feet are preserved for future processing.

\subsection{Feature Extraction}
The feature extraction process is the same as LOAM\cite{zhang2014loam}.  For each point $p_i$, its roughness value c is calculated by the mean of relative range differences between its neighbors $S$:

\begin{equation}
\begin{aligned}
c = \frac{1}{|S|\cdot ||r_i||}||\sum_{j\in S, j \neq i} (r_j - r_i)||
\end{aligned}
\end{equation}

Points that have a roughness value above a threshold $c_th$ are called edge features, while those below the threshold are called planar features. The range image is divided into 6 sub-images horizontally. For each sub-image, points with top roughness value in each row are selected as edge features, and those with the lowest roughness value are selected as planar features. Note that ground points are never selected as edge features.

\subsection{Motion Estimation}
Motion estimation is to compute the transformation between two scans. Same as LeGO-LOAM\cite{shan2018lego}, the transform is found by a two-step point-to-edge and point-to-plane feature matching. For a point $p_{k+1,i}$ in current frame, if $p_{k,j}$, $p_{k,l}$ are points on the edge line of previous frame, the point-to-edge distance is
\begin{equation}
    \begin{aligned}
    d_e = \frac{|(p_{k+1,i}-p_{k,j}) \times (p_{k+1,i}-p_{k,l})|}{p_{k,j} - p_{k,l}}
    \end{aligned}
\end{equation}
For a point $p_{k+1,i}$ in current frame, if $p_{k,j}$, $p_{k,l}$, $p_{k,m}$ are points on the planar plane of the previous frame, the point-to-plane distance is calculated by
\begin{equation}
    \begin{aligned}
    d_p = \frac{\begin{bmatrix}
    (p_{k+1,i}-p_{k,j}) \\
    (p_{k,j}-p_{k,l}) \times (p_{k,j}-p_{k,m})
    \end{bmatrix}}
    {(p_{k,j}-p_{k,l}) \times (p_{k,j}-p_{k,m})}
    \end{aligned}
\end{equation}

In the first step of the LM optimization, $[t_z, \theta_{roll}, \theta_{pitch}]$ are estimated by point-to-plane matching. Then the remaining $[t_x, t_y, \theta_{yaw}$ are estimated by point-to-line matching. This is more efficient than directly computing the six degrees of freedom.

\section{DATA and Platform}
\label{sec:data}
The framework proposed in this paper is validated using datasets gathered from Velodyne VLP-16 and HDL-64E 3D lidars. The VLP-16 measurement range is up to 100m with an accuracy of $\pm 3$cm. It has a vertical ﬁeld of view (FOV) of $30 ^ { \circ } \left( \pm 15 ^ { \circ } \right)$ and a horizontal FOV of $360 ^ { \circ }$ . The 16-channel sensor provides a vertical angular resolution of 2 ◦ . The horizontal angular resolution varies from $0.1 ^ { \circ }$ to $0.4 ^ { \circ }$ based on the rotation rate. And also with a scan rate of 10HZ which provides a horizontal angular resolution of $0.2 ^ { \circ }$. 

The UGV used in this paper is the Clearpath Jackal. Powered by a 270 Watt-hour Lithium battery, it has a maximum speed of 2.0m/s and a maximum payload of 20kg.

The LiDAR data is collected in both indoor and outdoor environments. The indoor case is a narrow and long corridor and a large lobby collected on the 1st floor of Newell-Simon Hall, and Field Robotics Center of Carnegie Mellon University. The outdoor case is collected in an environment that has two slopes of different gradients.

We also used another dataset collected by an autonomous driving ATV system which has RTK-GPS data as ground truth and can be used in algorithm comparison and evaluation. This data set is also collected by a Velodyne VLP-16 but mounted at a different angle on the vehicle which is not parallel to the ground.

\begin{figure}[h]
    \centering
    \includegraphics[width=0.8\linewidth]{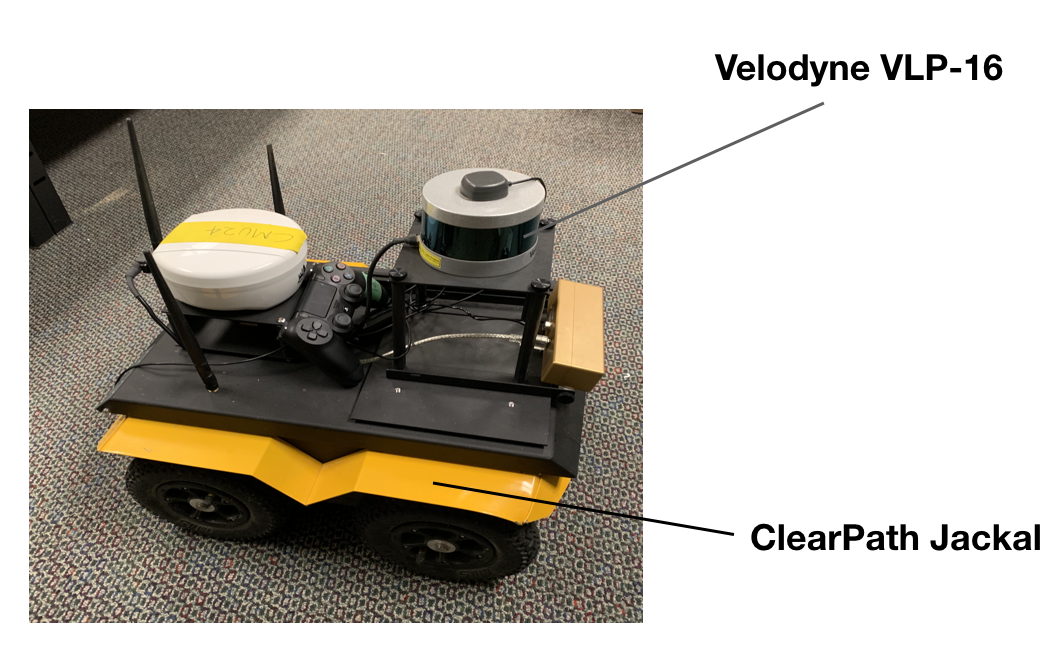}
    \caption{ClearPath Jackal Platform}
    \label{fig:exp1}
\end{figure}

\section{IMPLEMENTATION}
\label{sec:implement}
\subsection{Refined Ground Point Extraction}
We use breath-first-search to do point clustering. Starting with a point in the nearest row at the front, new points are added to this cluster if the vertical angles between neighboring points are below a threshold. We manually set the threshold for a vertical angle to 10 degrees and the threshold for cluster size to 100 points. These thresholds can be modified for different sensors.

\subsection{Pipeline}
\begin{figure}[h]
    \centering
    \includegraphics[width=0.9\linewidth]{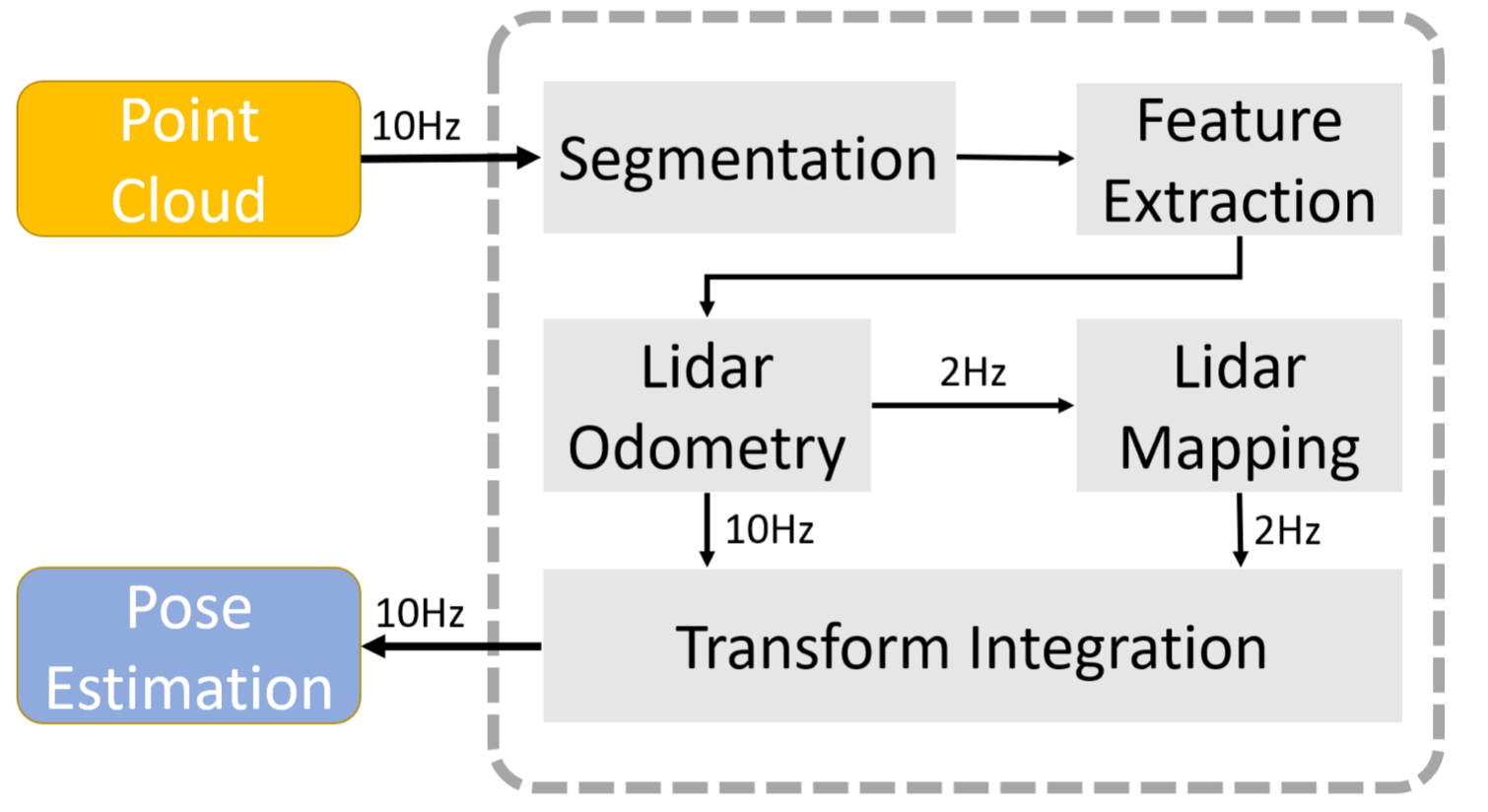}
    \caption{System Overview}
    \label{fig:overview}
\end{figure}

We adopted the similar pipeline as LeGO-LOAM~\cite{shan2018lego}, whose system overview is shown in Figure~\ref{fig:overview}. All the modules are implementes in C++ and on ROS.

The input of this pipeline is a point cloud map. The point cloud map is represented as a 16*1800 range map with each element being the distance between the center of the sensor and the point.
The first step is point segmentation using the method described in section~\ref{sec:methods}. Now the range map has been added another element: clustering label. The second step is feature extraction based on the range and clustering results. Then frame-to-frame registration is performed based on extracted features and previous frames. In the meanwhile, a low-frequency mapping on the global pointclouds is performed. Finally, the global mapping and frame-to-frame registration results are fused to get the predicted transformation.

\section{EXPERIMENTS AND RESULTS}
\label{sec:experiments}
We now describe a series of experiments to qualitatively and quantitatively analyze two methods, LeGO-LOAM with new ground extraction and original LeGO-LOAM.

\subsection{Jackal Mobile platform}
    
We manually drive the robot in both indoor and outdoor environments. We first show a qualitative comparison of extracted ground surface between our method and the original LeGO LOAM.
    \begin{figure}[th]
    \centering
    \begin{subfigure}[b]{0.4\textwidth}
        \includegraphics[width=\textwidth]{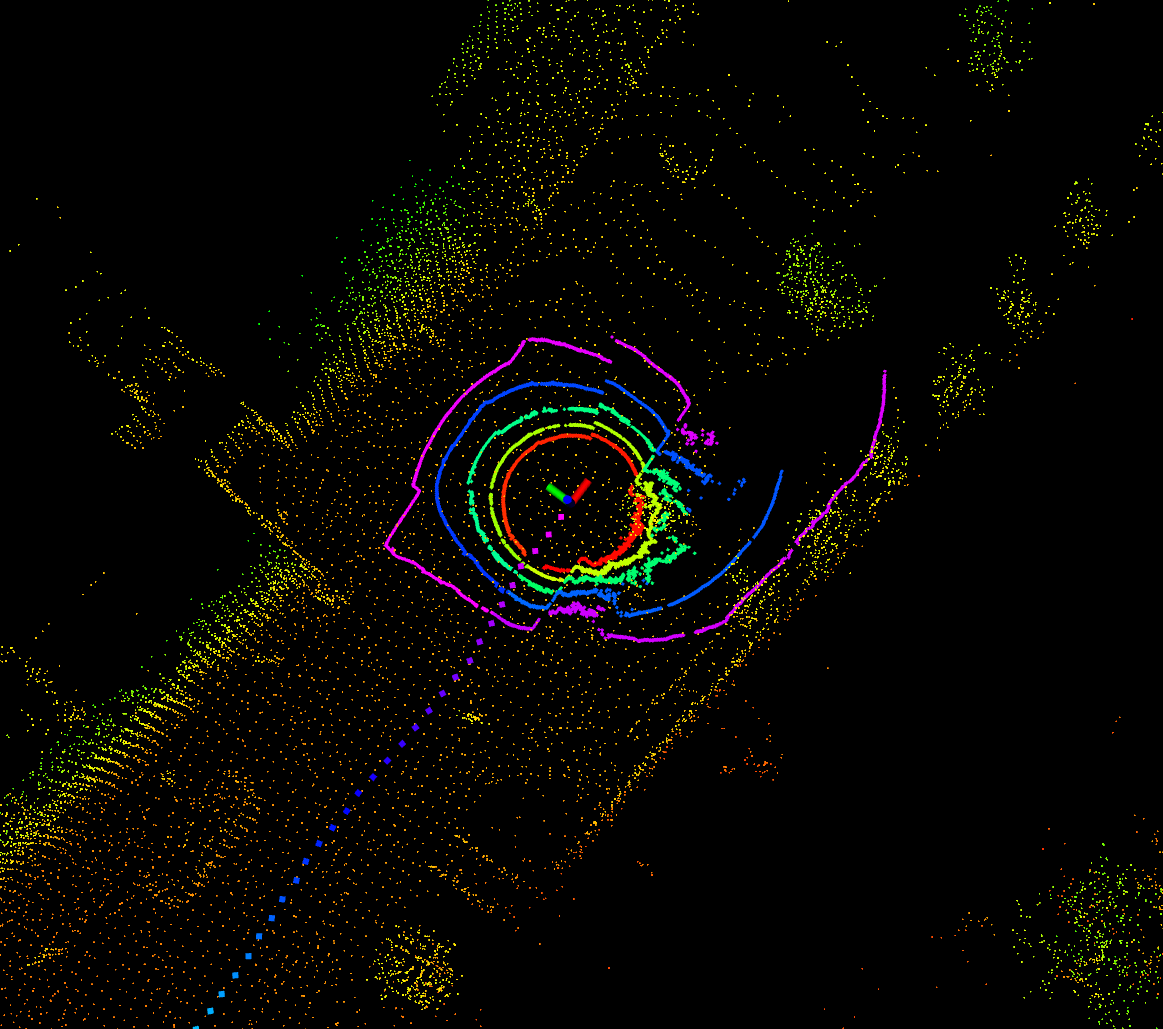}
        \caption{Noise LeGO LOAM}
    \end{subfigure}
    ~ %add desired spacing between images, e. g. ~, \quad, \qquad, 
    \begin{subfigure}[b]{0.4\textwidth}
        \includegraphics[width=\textwidth]{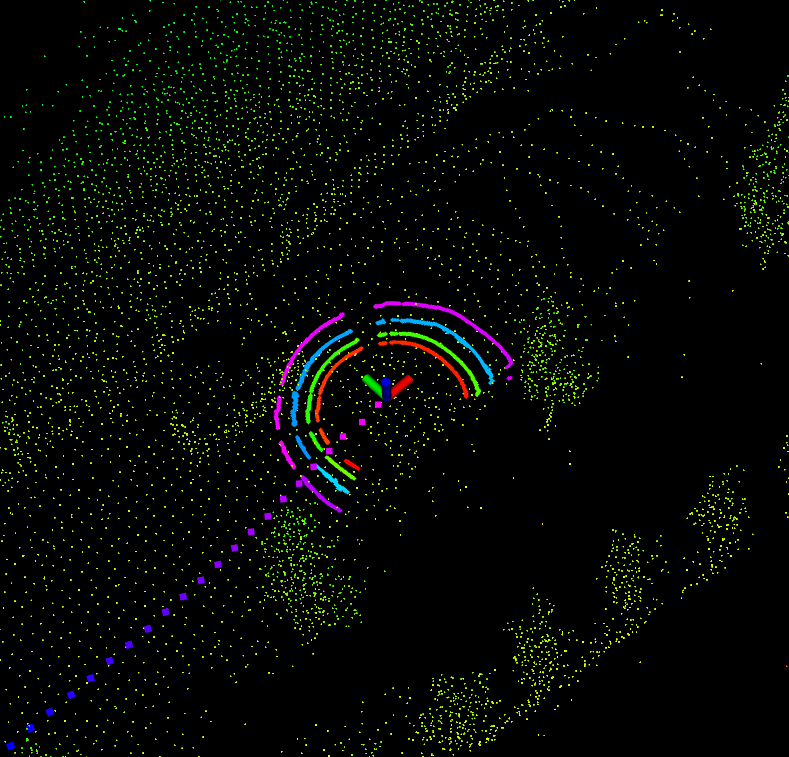}
        \caption{Noise Ground Refined}
    \end{subfigure}
\caption{Ground Surface Visualization}
\label{fig:exp2}
\end{figure}

As shown in Figure~\ref{fig:exp2}, the noise of the ground surface extracted from the original LeGo-LOAM has much more noise than the ground planar extracted from our method.

The reason is that LeGo-LOAM directly uses the ground removal method from LOAM, which aims to remove ground as well as other noise around the ground, to extract ground planar and use it for registration. If we want to leverage the ground planar for registration, the extracted ground needs to be clean and consistent.

Actually, the LeGo-LOAM fails in this outdoor data set sometimes during the test. The localization fails especially when the robot is turning. But using our method, there is no failure case that happened during our test.

\begin{figure}[th]
    \centering
    \begin{subfigure}[b]{0.4\textwidth}
        \includegraphics[width=\textwidth]{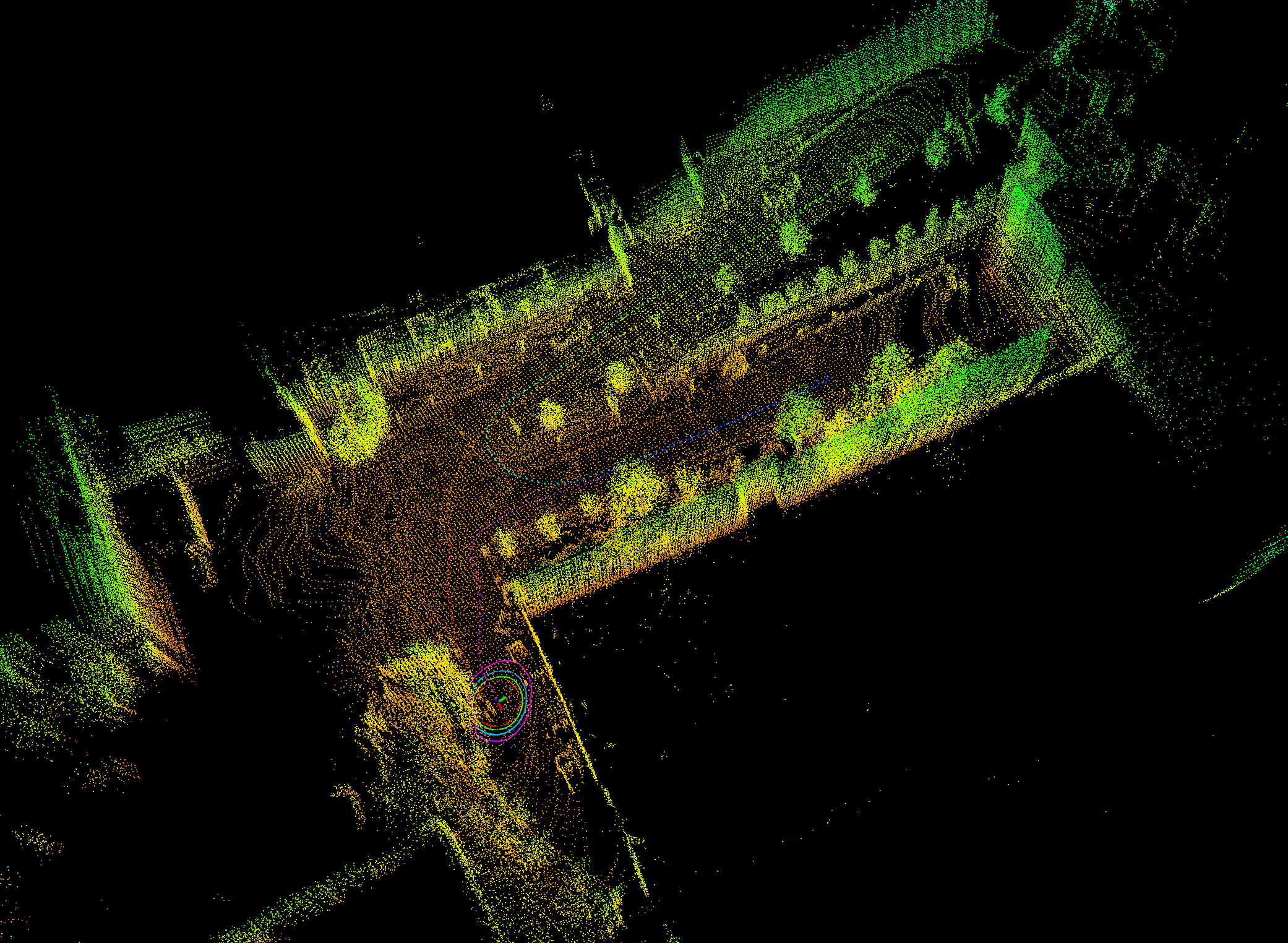}
        \caption{Outdoor Wean Hall}
    \end{subfigure}
    ~ %add desired spacing between images, e. g. ~, \quad, \qquad, 
    \begin{subfigure}[b]{0.4\textwidth}
        \includegraphics[width=\textwidth]{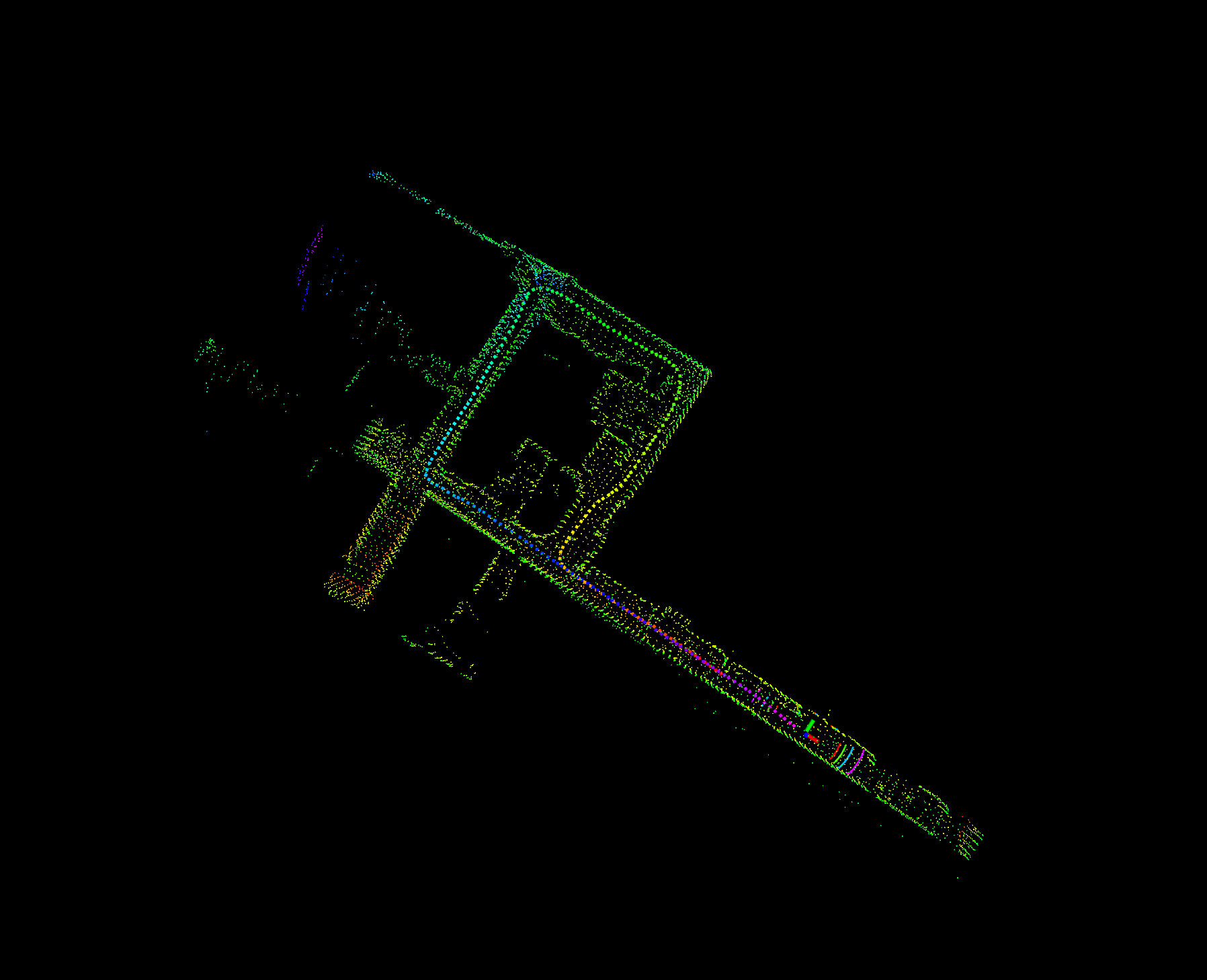}
        \caption{NSH 1st Floor}
    \end{subfigure}
    ~ %add desired spacing between images, e. g. ~, \quad, \qquad, 
    \begin{subfigure}[b]{0.4\textwidth}
        \includegraphics[width=\textwidth]{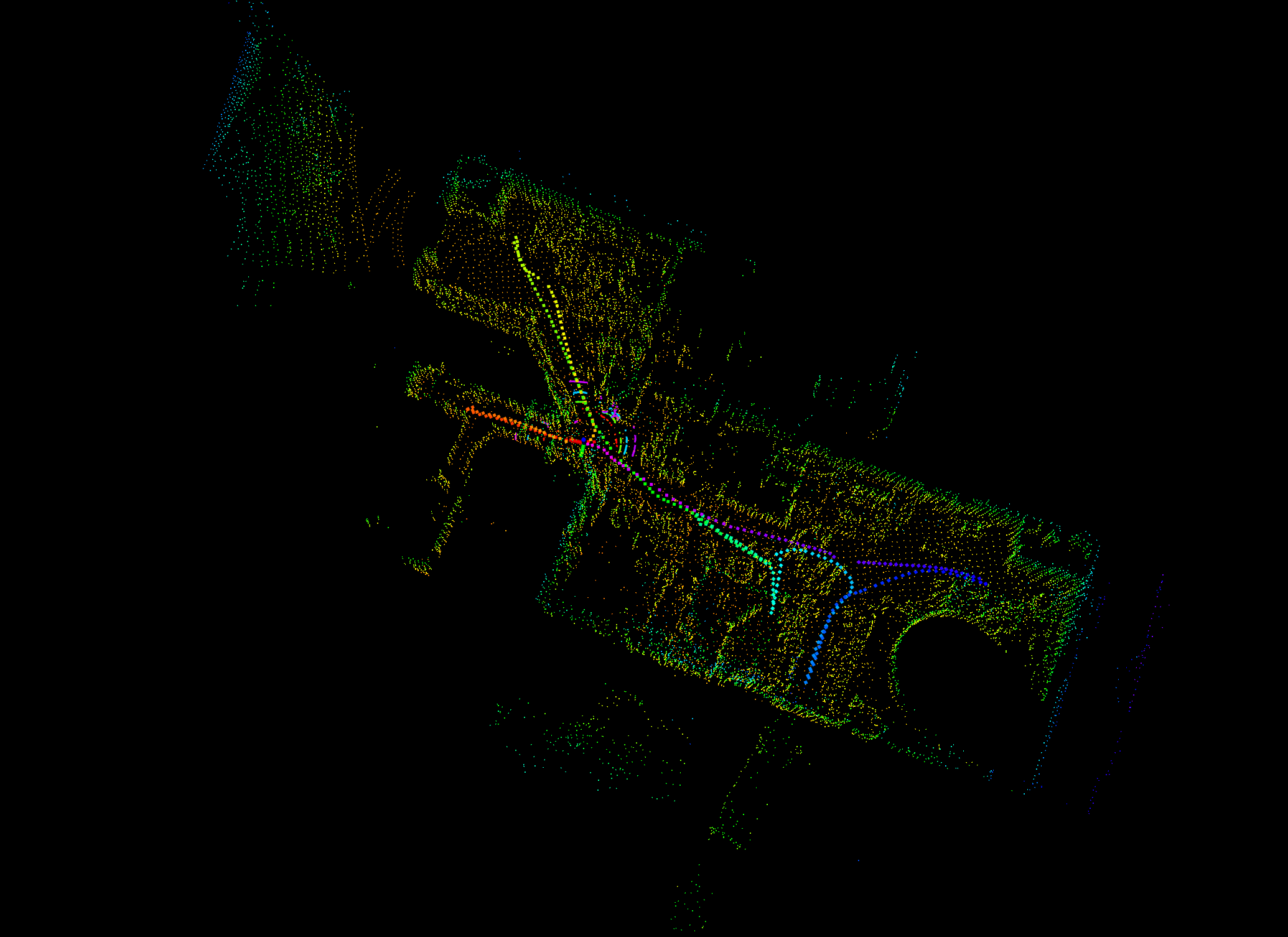}
        \caption{High Bay Lab}
    \end{subfigure}
\caption{Ground Refined SLAM Results}
\label{fig:exp3}
\end{figure}

Figure~\ref{fig:exp3} are some visualized results from our collected data set.

We also recorded video of our results (\href{https://drive.google.com/file/d/162MVT3BIkQZI4Tyy9ulab_TYwCAo-8nF/view}{link}) and video of LeGo-LOAM (\href{https://drive.google.com/file/d/19ctJxPRJx4_w5BlpdT06QSv2Yp2Ft0HB/view}{link}) on our collected data. The comparison of noise can be view in (\href{https://drive.google.com/open?id=1Bpwgv3EY1MpPJdiJDNshLSv4Si26Jvzk}{link}).

\subsection{ Autonomous driving ATV system}
    
We also evaluate our algorithm in a data set collected by an autonomous driving ATV system, as shown in Figure~\ref{fig:exp1}. It has ground truth collected by the RTK-GPS system, which has localization accuracy within 50cm. The result below has ground truth, LeGO LOAM result, and our method result is shown in blue, yellow, red respectively. 
\begin{figure}[th]
    \centering
    \includegraphics[width=0.9\linewidth]{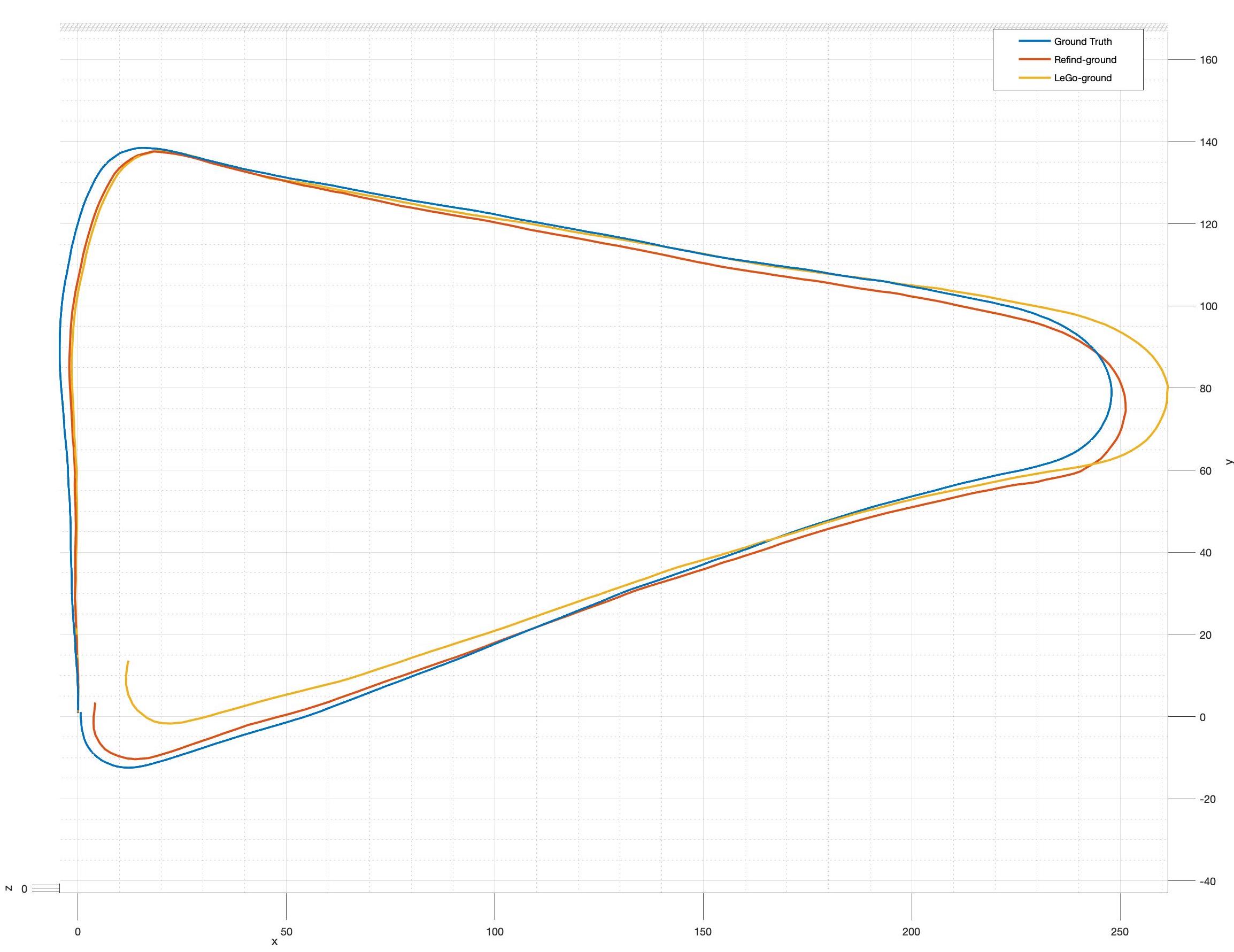}
    \caption{ATV data set}
    \label{fig:exp1}
\end{figure}

Table~\ref{tab} shows the result of final drift and comparation result running ATV lidar data. The data is collected by a HDL-64E LiDAR, which also has a horizontal FOV of $360 ^ { \circ }$ but 48 more channels compared to Velodyne VLP-16. The vertical FOV of the HDL-64E is $26.9 ^ { \circ }$.

\begin{table}[th]
\begin{tabular}{l|l|l|l}
          & DISTANCE (m)   & DRIFT & PERCENTAGE \\ \hline \hline
LeGo-LOAM & 669.930      & 18.354    & 2.739$\%$            \\ \hline
OURS      & 669.930       &  6.367  & 0.950$\%$          
\end{tabular}
\caption{Drift result of ATV dataset}
\label{tab}
\end{table}

It is shown in both figures and the drift table that our method outperforms the original LeGo-LOAM~\cite{shan2018lego} algorithm.

\section{CONCLUSION AND FUTURE WORK}
\label{sec:conclusion}
After implementing our method of ground plane extraction, the results shown above indicate that less noisy ground point extraction does improve the performance of frame-to-frame registration and provide a better SLAM result. However, the algorithm we implemented in this project inherited from LeGo-LOAM~\cite{shan2018lego} still uses point-to-plane registration and point-to-point registration without adding plane-to-plane registration, which we believe could provide a better registration result in pitch and roll orientation as well as linear offset in the z-direction.

In the future, we also plan to integrate plane-to-plane registration into the global optimization process, which may provide a more accurate and robust performance.
\bibliographystyle{plain}
\bibliography{references}

\begin{thebibliography}{1}

\bibitem{pfrunder2017real}
Andreas Pfrunder, Paulo~VK Borges, Adrian~R Romero, Gavin Catt, and Alberto
  Elfes.
\newblock Real-time autonomous ground vehicle navigation in heterogeneous
  environments using a 3d lidar.
\newblock In {\em 2017 IEEE/RSJ International Conference on Intelligent Robots
  and Systems (IROS)}, pages 2601--2608. IEEE, 2017.

\bibitem{shan2018lego}
Tixiao Shan and Brendan Englot.
\newblock Lego-loam: Lightweight and ground-optimized lidar odometry and
  mapping on variable terrain.
\newblock In {\em 2018 IEEE/RSJ International Conference on Intelligent Robots
  and Systems (IROS)}, pages 4758--4765. IEEE, 2018.

\bibitem{tang2018learning}
Tim Tang, David Yoon, Fran{\c{c}}ois Pomerleau, and Timothy~D Barfoot.
\newblock Learning a bias correction for lidar-only motion estimation.
\newblock In {\em 2018 15th Conference on Computer and Robot Vision (CRV)},
  pages 166--173. IEEE, 2018.

\bibitem{zermas2017fast}
Dimitris Zermas, Izzat Izzat, and Nikolaos Papanikolopoulos.
\newblock Fast segmentation of 3d point clouds: A paradigm on lidar data for
  autonomous vehicle applications.
\newblock In {\em 2017 IEEE International Conference on Robotics and Automation
  (ICRA)}, pages 5067--5073. IEEE, 2017.

\bibitem{zhang2014loam}
Ji~Zhang and Sanjiv Singh.
\newblock Loam: Lidar odometry and mapping in real-time.
\newblock In {\em Robotics: Science and Systems}, volume~2, page~9, 2014.

\end{thebibliography}

\end{document}